# Detailed Aerial Mapping of Photovoltaic Power Plants Through Semantically Significant Keypoints

**Viktor Kozák** *[1,2], **Jan Chudoba** [2], **Libor Přeučil** [2]

[1] *Czech Technical University in Prague, Faculty of Electrical Engineering, Department of Cybernetics, Karlovo náměstí 13, Prague 2, 121 35, Czech Republic, viktor.kozak@cvut.cz*
[2] *Czech Technical University in Prague, Czech Institute of Informatics, Robotics, and Cybernetics, Jugoslávských partyzánů 1580/3, Prague 6, 160 00, Czech Republic, jan.chudoba@cvut.cz, libor.preucil@cvut.cz*

**Abstract**

An accurate and up-to-date model of a photovoltaic (PV) power plant is essential for its optimal operation and maintenance. However, such a model may not be easily available. This work introduces a novel approach for PV power plant mapping based on aerial overview images. It enables the automation of the mapping process while removing the reliance on third-party data. The presented mapping method takes advantage of the structural layout of the power plants to achieve detailed modeling down to the level of individual PV modules. The approach relies on visual segmentation of PV modules in overview images and the inference of structural information in each image, assigning modules to individual benches, rows, and columns. We identify visual keypoints related to the layout and use these to merge detections from multiple images while maintaining their structural integrity. The presented method was experimentally verified and evaluated on two different power plants. The final fusion of 3D positions and semantic structures results in a compact georeferenced model suitable for power plant maintenance.

## 1. Introduction

The growing adoption of solar energy has driven the need for efficient monitoring and maintenance of photovoltaic (PV) power plants. A power plant model suitable for long-term maintenance should include the positions of individual PV modules along with their orientations, types, and placement in the structure of the power plant. In order to support the deployment of autonomous aerial inspection procedures [1], positions should be aligned with a global coordinate system referenced by the global navigation satellite system (GNSS). A direct assignment of PV modules to known segments can also be used to plan regular or more detailed inspection flights above segments with known defects. The required model accuracy is generally application-dependent and may consider potential shifts in global coordinates and the internal consistency of the model.

A power plant model can be obtained through several methods. Ideally, a technical drawing with the original design should be available. Another option is to use high-quality satellite or orthophotographic images from public mapping services to infer the layout of the power plant. However, both of these options rely on available historical data and require verification against the current state of the power plant, as the layout can undergo changes during construction or due to continuous maintenance [2]. This verification entails either manual inspection or acquisition of new data. A different approach to power plant mapping can rely on direct acquisition and processing of aerial images. The benefit of this is the guaranteed consistency with the actual state of the power plant.

We present an approach for detailed and efficient mapping of PV power plants using aerial overview images. The images are used to create a raw 3D model of the surveyed area. Subsequently, we detect individual PV modules and determine the layout of the power plant in each image. However, the accuracy of the 3D reconstruction is limited, which makes direct matching of PV module detections between individual images unfeasible. The main contribution of our work is the use of visually identifiable keypoints related to the structural layout that are used to merge the detected structures from overview images taken from different viewpoints. This allows us to map power plants with a lesser number of images compared to other approaches.

The final step fuses the 3D positions and the semantic structures. This results in a compact georeferenced model suitable for operation, maintenance, and autonomous inspection.

## 2. Related work

The use of unmanned aerial systems has become increasingly popular in various applications, including PV power plant inspection [3], [4], maintenance [2], and site selection for new PV installations [5]. A major focus is placed on computer vision techniques for autonomous detection and localization of solar power plant installations and PV modules.

Visual segmentation in satellite-based images was addressed in [6]. The authors published a large multiresolution dataset and tested several segmentation methods. However, the focus was limited to the segmentation of the power plant area without inferring its exact layout. In [7], a geographic information system and a manually created power plant plan in vector format were used. In [8], an edge detection method was used to infer the layout of a power plant on publicly available high-altitude orthophotos. However, manual correction of the detected results was necessary. The power plant layout was subsequently combined with publicly available elevation maps. The work also compares the performance of three different PV module detection methods.

Detailed mapping using video sequences from flights along rows of PV bench installations was addressed in [9]. The authors selected a subset of keyframes based on their global positioning system coordinates and visual overlap and used structure-from-motion (SfM) technique to create a 3D reconstruction of the power plant. The work used Mask R-CNN [10] for PV module segmentation in thermal images. The downside of this approach is the reliance on module tracking in videos captured from low-altitude flights. Although the authors verified the increasing module throughput when processing two or three bench rows simultaneously.

The authors of [11] used SfM and multi-view stereo techniques to create terrain and surface models of the power plant. Subsequently, true orthoimages were generated. A similar approach was adopted in [12]. In both works, the detection of PV modules was performed on the orthoimages by applying different thresholding and contour detection operations. Density-based clustering was used to group the detected PV modules into arrays.

An additional repair procedure was introduced in [12], where undetected PV modules are added based on the expected geometry. In contrast to this, our mapping pipeline relies on redundancy in image detections and maps only the modules that were visually confirmed.

Existing aerial mapping methods [9,11,12] require hundreds of aerial images. In contrast, our approach reduces the number of required images by an order of magnitude; thus, significantly reducing the necessary flight and computation times. By leveraging structural design patterns commonly found in a large number of PV power plants, our method is able to produce a detailed model with module-level accuracy and clear bench boundaries. While the method is broadly applicable, its reliance on the general design concepts can limit its applicability in irregular layouts.

## 3. Methodology

The presented approach is based on the use of sparse overview images. The mapping pipeline (Figure 1) can be divided into five main steps.

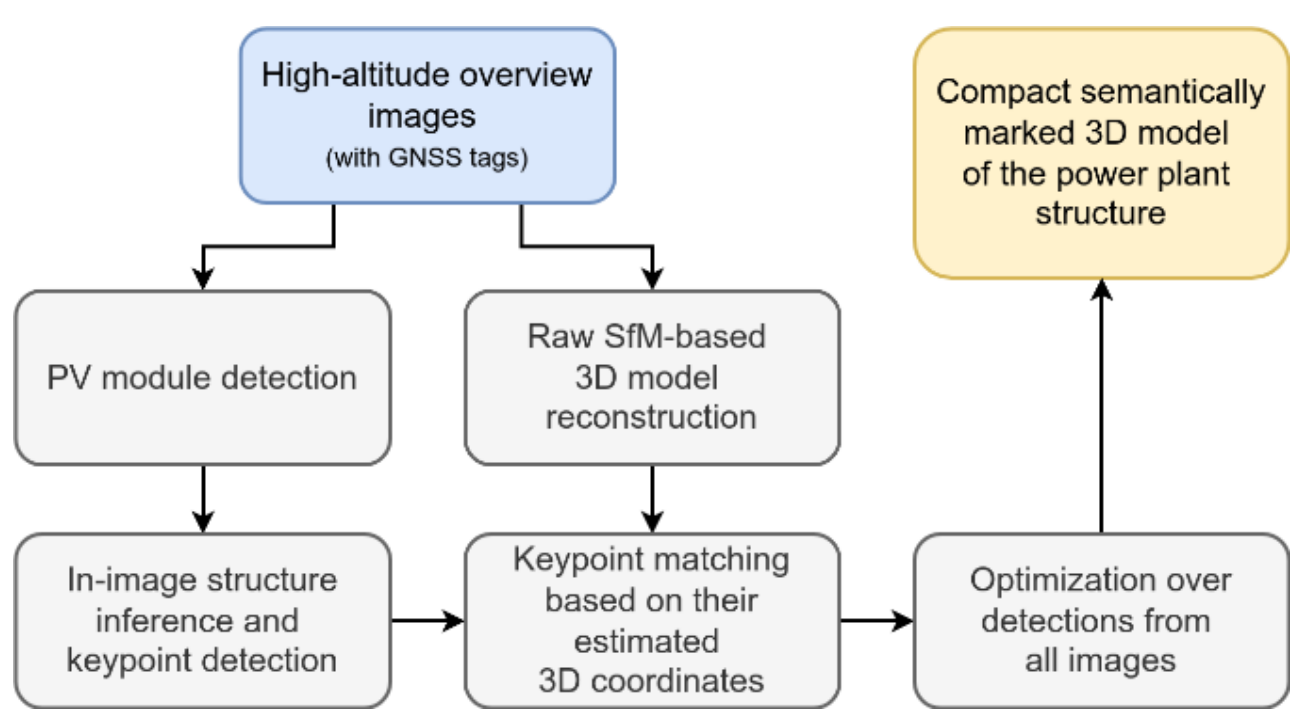

**Figure 1.** PV power plant mapping pipeline.

First, individual PV modules are detected in the overview images. Then, the layout of the detected modules is determined in each image, assigning them to individual rows, columns, and benches (Figure 2). Semantically significant visual keypoints are identified within these structures.

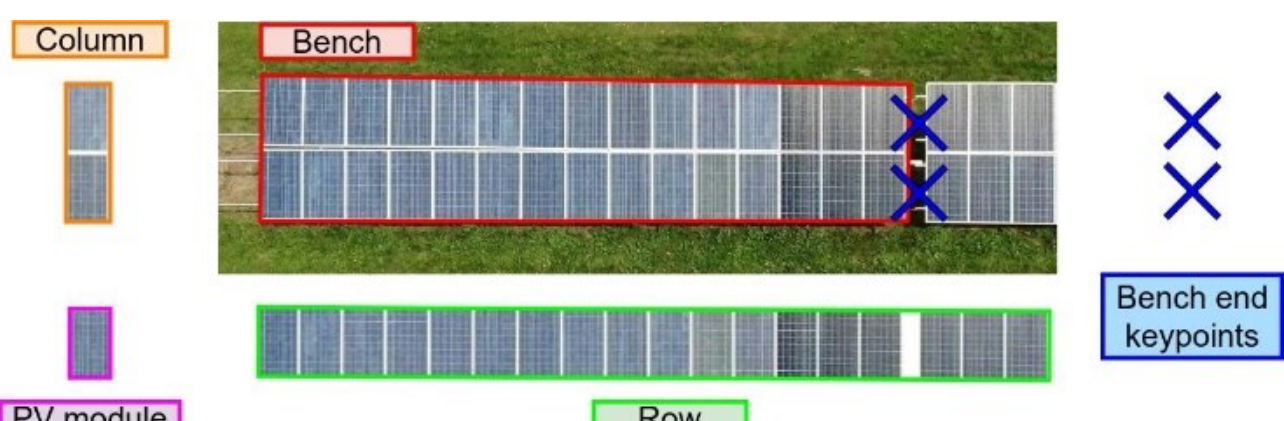

**Figure 2.** A depiction of individual power plant elements.

A raw 3D model is reconstructed from the overview images using the SfM technique, resulting in a point cloud that includes point positions, colors, and normals. The estimated camera positions of individual frames are also refined during the procedure.

The 3D model is used to estimate the 3D coordinates of each keypoint detection. These are used to pair keypoints between individual images and determine the complete structure of the power plant. From this we obtain information on the allocation of each detected PV module within the global power plant structure.

Finally, coordinates of each PV module are calculated by merging their estimated positions from all detection instances. These can be further optimized using conventional design criteria for power plant layouts.

### 3.1. Data collection and 3D model reconstruction

The process starts with the acquisition of power plant images captured during an overview flight using a drone equipped with a high-resolution camera. These images should be marked with GNSS tags for subsequent georeferencing and cover the entire area of the power plant. Depending on the equipment used, altitudes of 60-100 meters above ground provide a good balance between image resolution and coverage for PV module segmentation. To use the SfM technique, images should have sufficient overlaps (over 50 % is recommended).

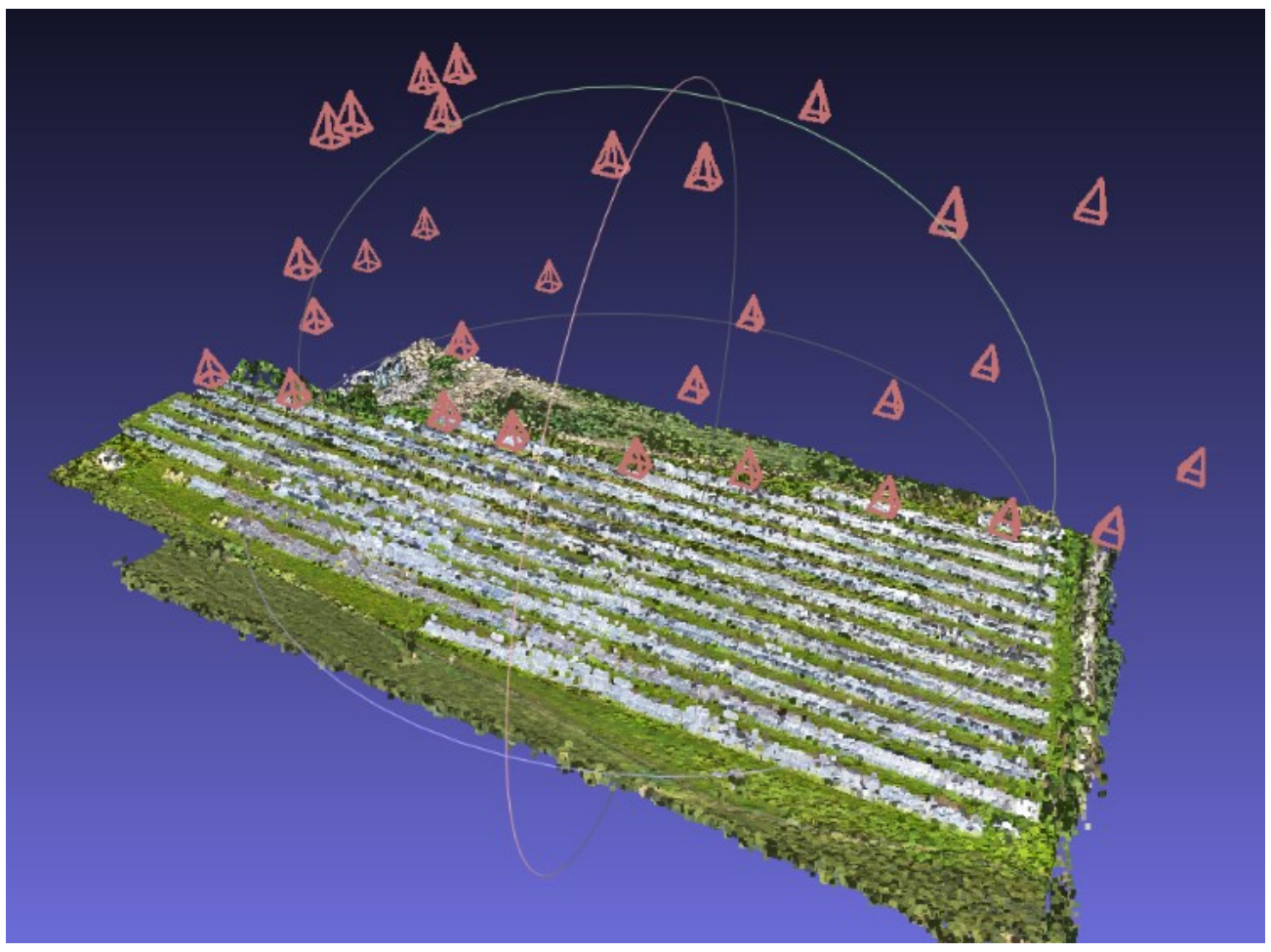

**Figure 3.** OpenSfM-generated 3D model and visualized camera positions of individual image frames used for the reconstruction.

We used the OpenSfM tool [13] to generate a raw 3D model of the power plant, an example of such model can be seen in Figure 3. The output of this tool includes a 3D model of the mapped environment, positions and rotations of individual frames, and images without distortions caused by the camera model. Camera parameters necessary for accurate image processing and 3D reconstruction [14] are estimated by this tool during the process. It is also possible to use alternative 3D reconstruction frameworks, as long as these provide similar outputs.

The undistorted images are used for the detection of PV modules. The 3D model and camera positions are used to estimate the 3D coordinates of the detected keypoints, which are used during the keypoint matching phase.

### 3.2. Visual segmentation of PV modules

In this work, we utilize the U-Net [15] and YOLOv8 [16] segmentation methods and assess their suitability for RGB images captured from higher altitudes.

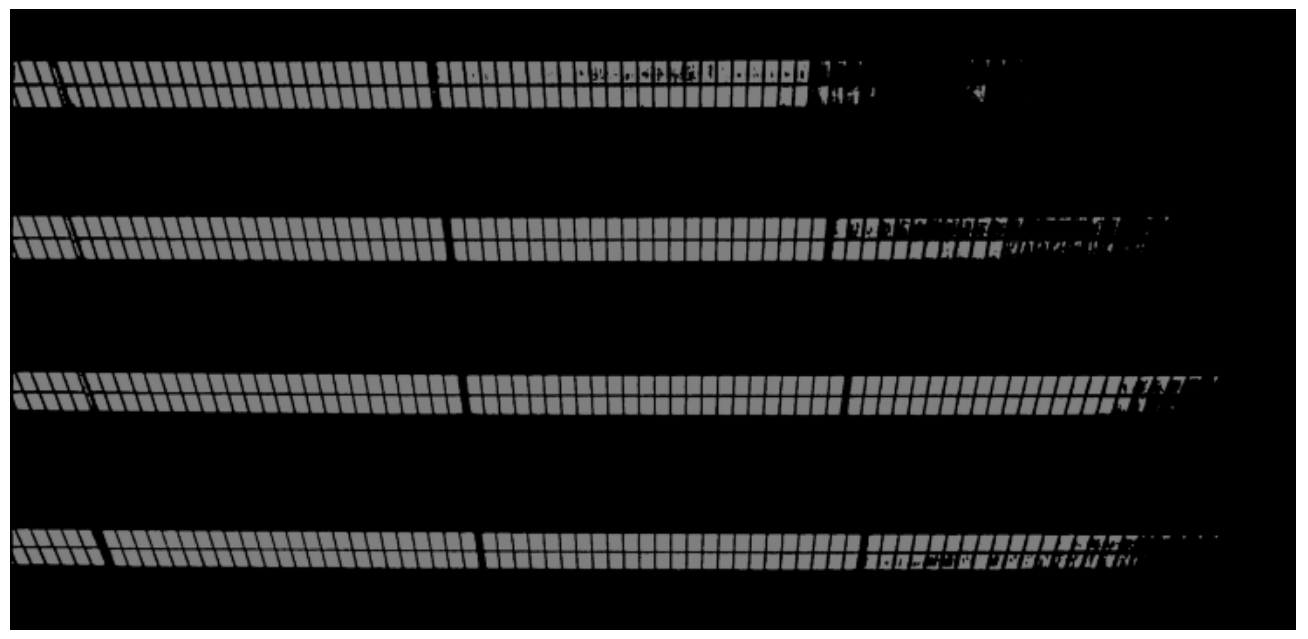

**Figure 4.** U-Net segmentation mask.

The U-net segmentation model is easily scalable to larger images and works without significant adjustments when used on overview images. U-Net is used to create binary segmentation masks. An example of a U-Net generated segmentation mask can be seen in Figure 4. Morphological operations (erosion and dilation) are applied to these masks to effectively filter out noise and potential unwanted detections. Individual segments in the masks are separated and used to determine oriented bounding boxes defining the position and boundaries of each PV module. The method performed reasonably well; however, it encountered problems with several segments of the PV installation that were oversaturated due to sun reflections.

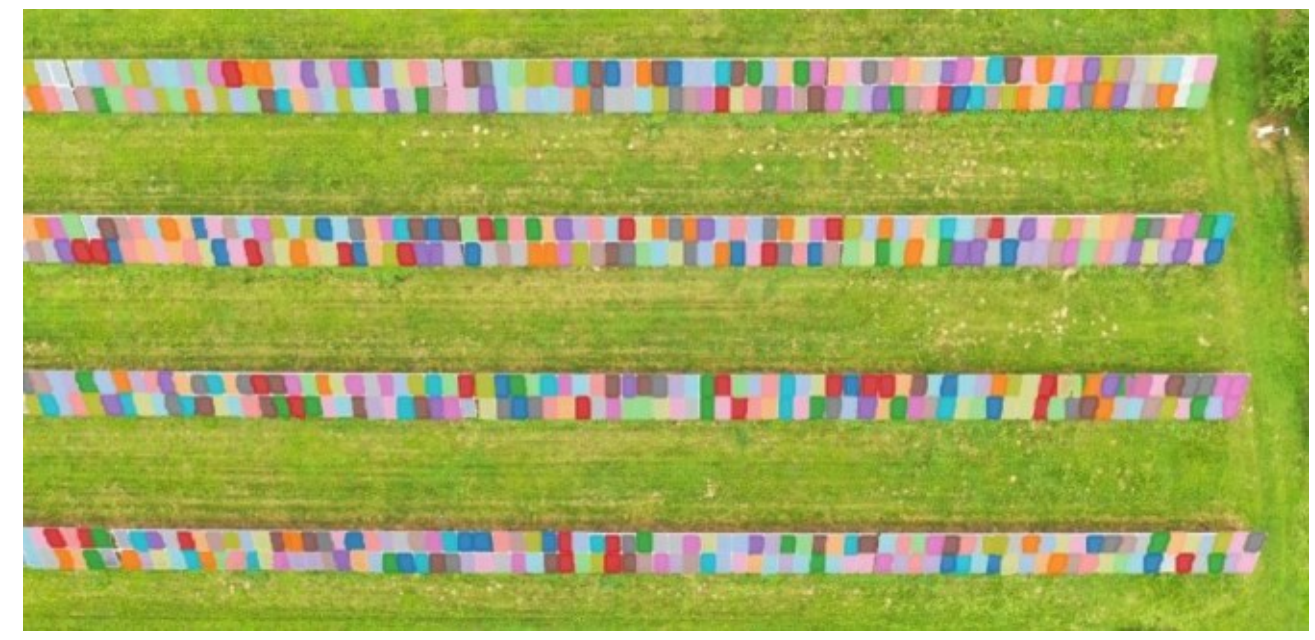

**Figure 5.** YOLO instance segmentation masks.

An example of YOLO-generated segmentation masks can be seen in Figure 5. YOLO-based instance segmentation detected a higher percentage of PV modules. However, it had trouble with the detection of PV modules between and at the end of the benches, which are crucial for the proposed mapping framework. Compared to U-Net, the positions of the detected masks and bounding boxes exhibited lower precision, with noticeable inconsistencies and shifts from the actual positions. The YOLO results were improved by post-processing. Individual segmentation masks were eroded, dilated, and used to create a new set of bounding boxes.

To leverage the benefits of both methods, a fusion of U-Net and YOLO detections was used. Since U-Net detections exhibit higher precision and stability, these are used as the primary foundation and then complemented with YOLO detections.

Detections on the edges of the image are discarded to remove incomplete PV modules. Detections with more than a 20 % overlap are also discarded. The median-sized bounding box is selected as a representative detection, and the results are filtered using the dimensions of the representative bounding box.

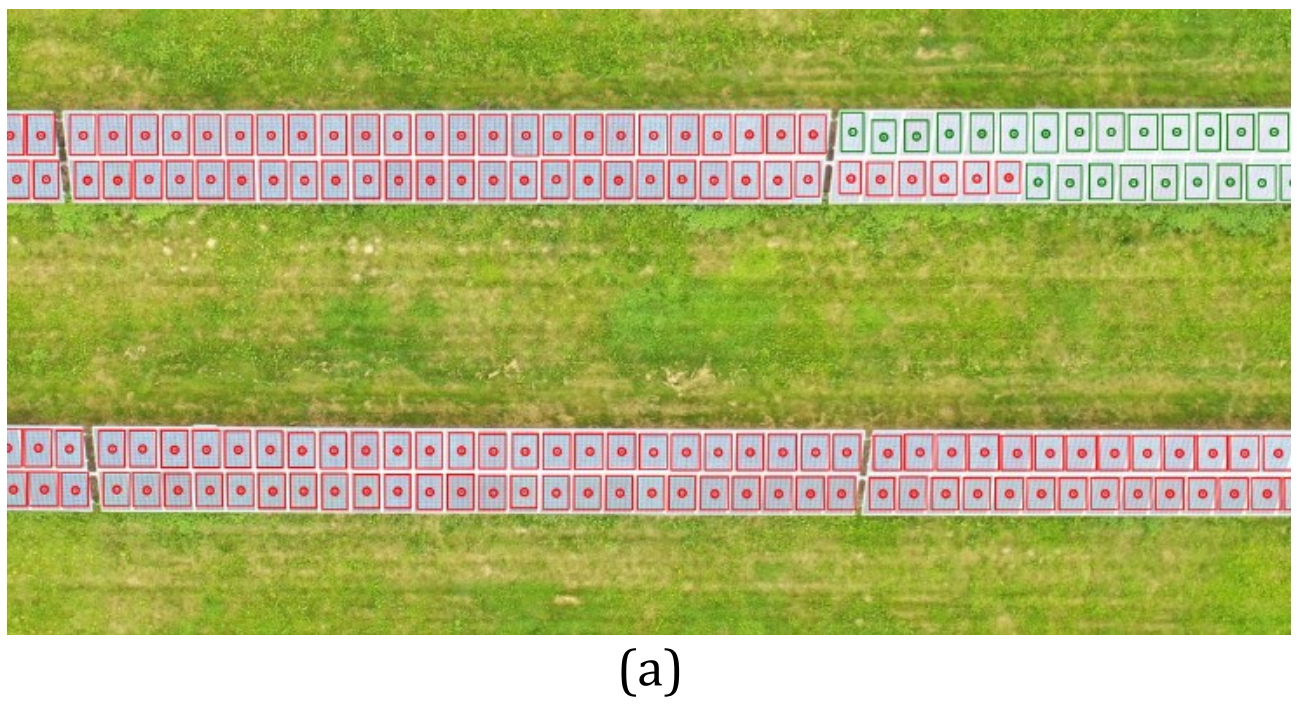

(a)

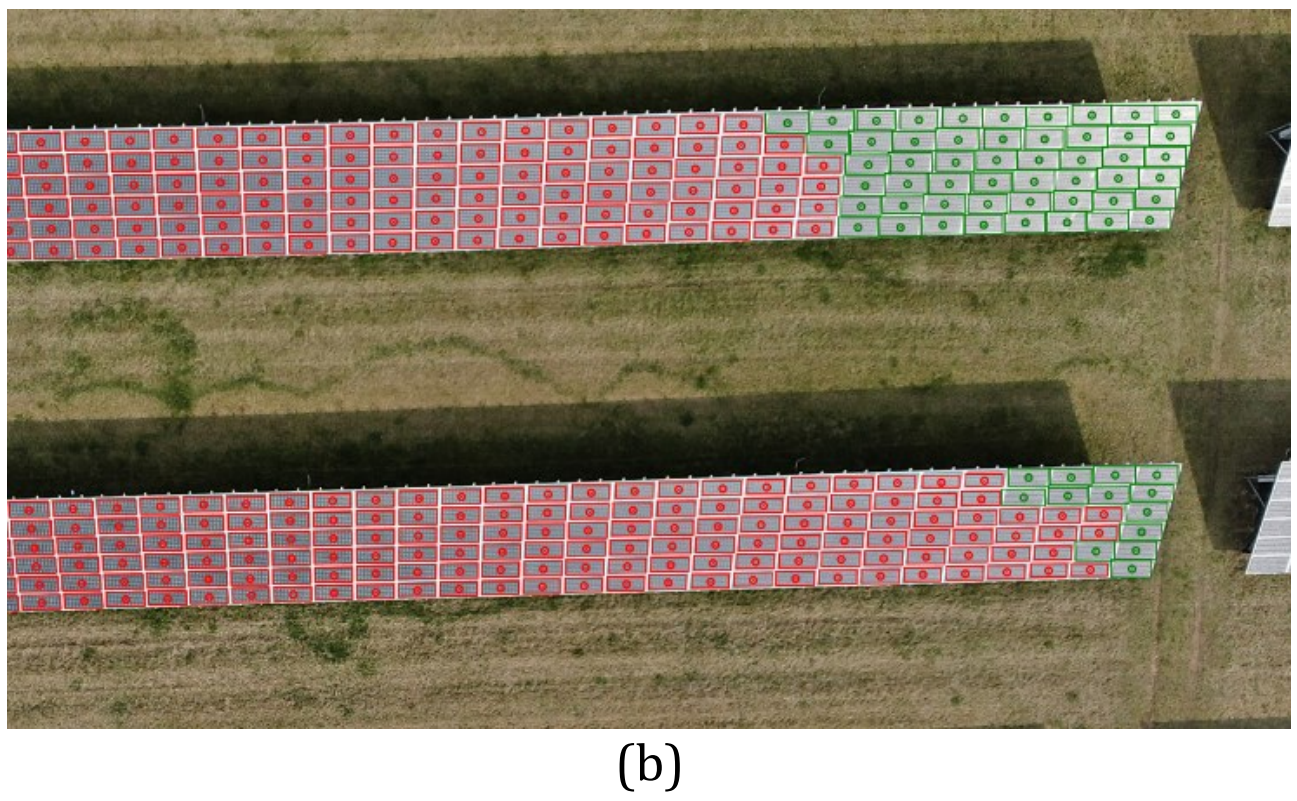

(b)

**Figure 6.** PV module centers and bounding boxes from U-Net based segmentation (red) complemented by YOLO detections (green). (a) Power plant 1. (b) Power plant 2.

These dimensions are also used to filter out YOLO detections on the basis of their distance from existing U-Net detections. The remaining YOLO detections are added to the final set of detected PV modules, significantly increasing the detection success rate. A visualization showing the fusion of detections from both methods can be seen in Figure 6. The output of this stage is a set of bounding boxes that define the in-image positions and dimensions of the detected modules.

### 3.3. In-image semantic structure inference

Bounding box centers, representing individual PV module detections, are used to infer the power plant layout in each of the images. The method is designed for structured power plants with modules organized in benches, where each bench has a known number of rows. The RANSAC method is used to calculate line models that represent individual rows and identify their inlying bounding box centers (as shown in Figure 7). The scikit-learn implementation [17] is used. The maximum residual threshold value for RANSAC is set to 25 % of the height of the representative bounding box.

Rows belonging to the same bench are grouped. The Hausdorff distance is used to measure the distances between two lines $L$ and $K$ representing the rows in the image using Equation 1:

$$H(L, K) = \max(d(l_0, K), d(l_1, K), d(k_0, L), d(k_1, L)), \quad (1)$$

where $d(l_i, K)$ is the Euclidean distance between a border point $l_i$ of line $L$ to line $K$. The border point is the intersection between the line and the image border.

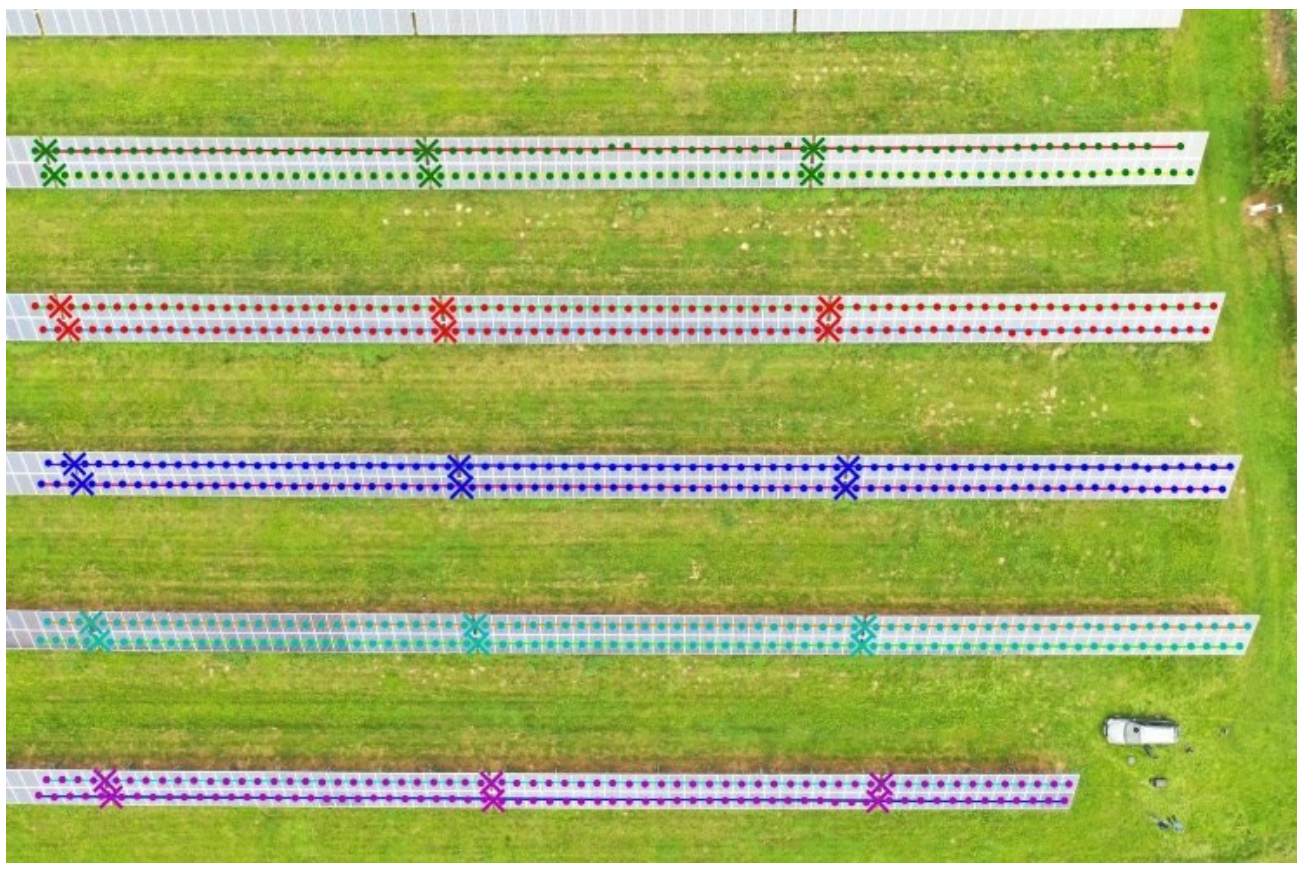

**Figure 7.** RANSAC-based module alignment and keypoint detection. Module centers are color-coded according to their respective benches. Crosses represent the detected bench gaps.

Lines are determined as coinciding if their maximal distance does not exceed twice the height of the representative bounding box. The validity of line inference and compounding is automatically verified using the known number of rows in each bench. Groups without the full number of inlying rows are discarded, since the in-bench (top-down) position of the rows cannot be determined.

We continue by identifying visually significant keypoints, specifically the gaps between the benches. First, the distances between the neighboring PV module detections are computed. We use the median distance $d_{med}$ and set an acceptable threshold for the expected distance between modules to 10 % ($th$ = 0.1). These are used to identify gap candidates based on the distance between neighboring modules $i$ and $j$ using Equation 2:

$$(1 + th) \cdot d_{med} < d_{ij} < 2(1 - th) \cdot d_{med} \quad (2)$$

The resulting keypoint is defined by the two neighboring modules. To verify the credibility of the gap candidates, we adopt a method described in [8]. We extract image patches that contain transitions between the modules and use histogram similarity to compare these. An illustration of the extracted image patches is shown in Figure 8. Bench gaps are identified by the presence of significantly darker transition patches. This method was chosen because it does not require any training data. In practice, alternative classification techniques could also be employed.

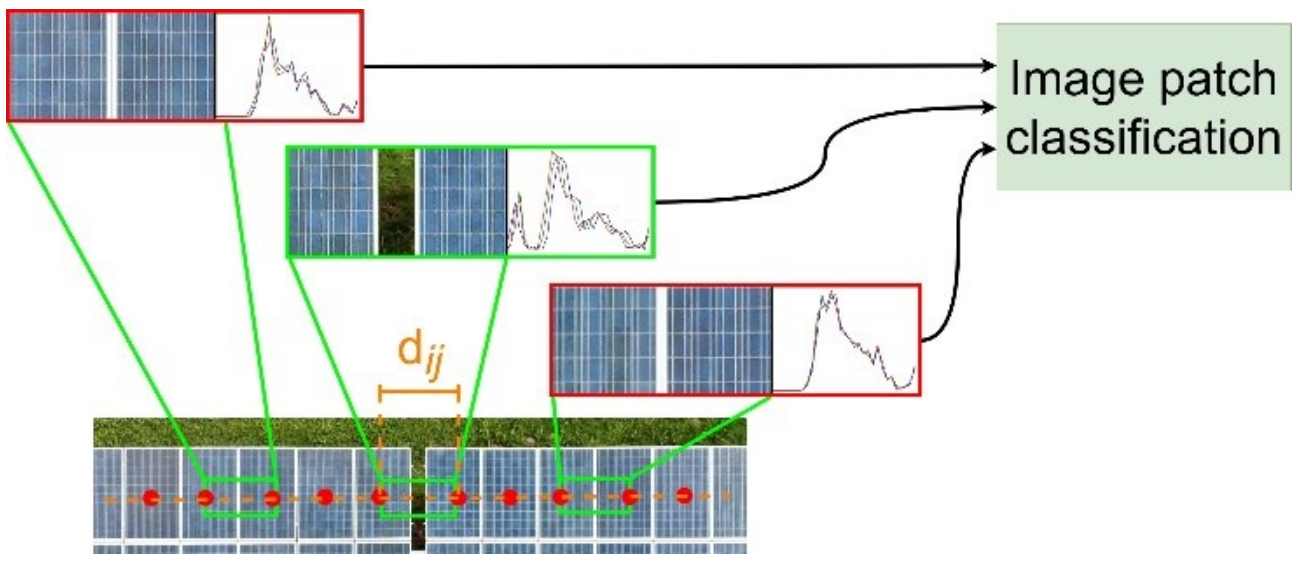

**Figure 8.** Illustration of extracted image patches between neighboring modules and their corresponding color histograms.

Given the limitations of the PV module segmentation methods, they are not guaranteed to detect all modules.

However, aside from the keypoint-defining modules, it is sufficient if a module is detected in at least one image. It is also possible to display the results of this phase (as shown in Figure 7) and manually assign or delete keypoints and modules if necessary.

Larger gaps in the sorted in-row sequence are identified as potential missing detections. We estimate their number and verify their credibility using the median distance (Equation 3):

$$(1-th)\cdot d_{med}\cdot(n+1) < d_{ij} < (1+th)\cdot d_{med}\cdot(n+1), \quad n \in \mathrm{N},\ n \le 8 \tag{3}$$

The $n$ hypothesized detections are placed evenly in the gap and confirmed as valid during the structure matching between individual images. We limit the maximum number of neighboring hypothesized detections ($n \le 8$).

The output is a structured set of row, keypoint, and PV module detections (as illustrated in Figure 9), each retaining information on their precise position in the image. Rows are grouped by their respective benches and sorted by their in-bench positions from north to south (from top to bottom of the image). These are further divided into sectors using the bench gap keypoints ($K^i$). PV modules ($m^{si}$) are assigned to individual rows and sectors and ordered by their position in the sector.

### 3.4. Keypoint matching

The accuracy of the raw 3D reconstruction model is limited. The raw model exhibits spatial discrepancies that are generally moderate but can occasionally exceed 0.6 meters. When the discrepancies exceed half the size of a PV module, the estimated positions cannot be confidently assigned to a specific module, which makes direct merging based on their estimated 3D positions impossible. Therefore, we propose the utilization of keypoints with lower density and greater mutual distance for merging. To this end, the bench gaps registered during the structure inference phase are used. Their 3D coordinates are calculated using the 3D model from the SfM-based reconstruction and used to match their detections between individual images.

The OpenSfM reconstruction output includes undistorted images and their respective refined camera positions. Since the camera positions and parameters are known, we can project a directional vector into the 3D space for each detected keypoint. By finding an intersection with the reconstructed 3D model, we are able to estimate a 3D position and normal for each detection. The 3D model is given as a point cloud, with each point containing information about its position, color, and normal. The position of the keypoint can be obtained by selecting the point with the minimum distance from the projected vector. We further improve on this by averaging the positions and normals of a small set of nearest points, increasing the robustness of the estimation.

Iterating over the set of all bench gaps detected in individual images, we group these using their mutual distance. We assign each keypoint to an existing group if its distance to at least one of the keypoints in the group is lower than a set threshold. The keypoint groups are further re-matched using the in-bench (top-down) position of their respective rows.

### 3.5. Fusion of the detected structures

The fusion of the detected structures between two images is illustrated in Figure 10. All keypoints matched in a single group belong to the same row in the global structure of the power plant. The same holds for each of the keypoints that lie on the same lines in their respective images. This is used to update a global line ID for each detected line and all of its associated PV module detections. The subsequent operations are performed separately for each global line ID.

Each sector is connected to at least one keypoint. The keypoints are connected either from the right or left side of the sector. We iterate through each set of matched keypoints and group sectors on one of the sides together. The same is done for sectors on the other side. This results in a set of groups of matched sectors.

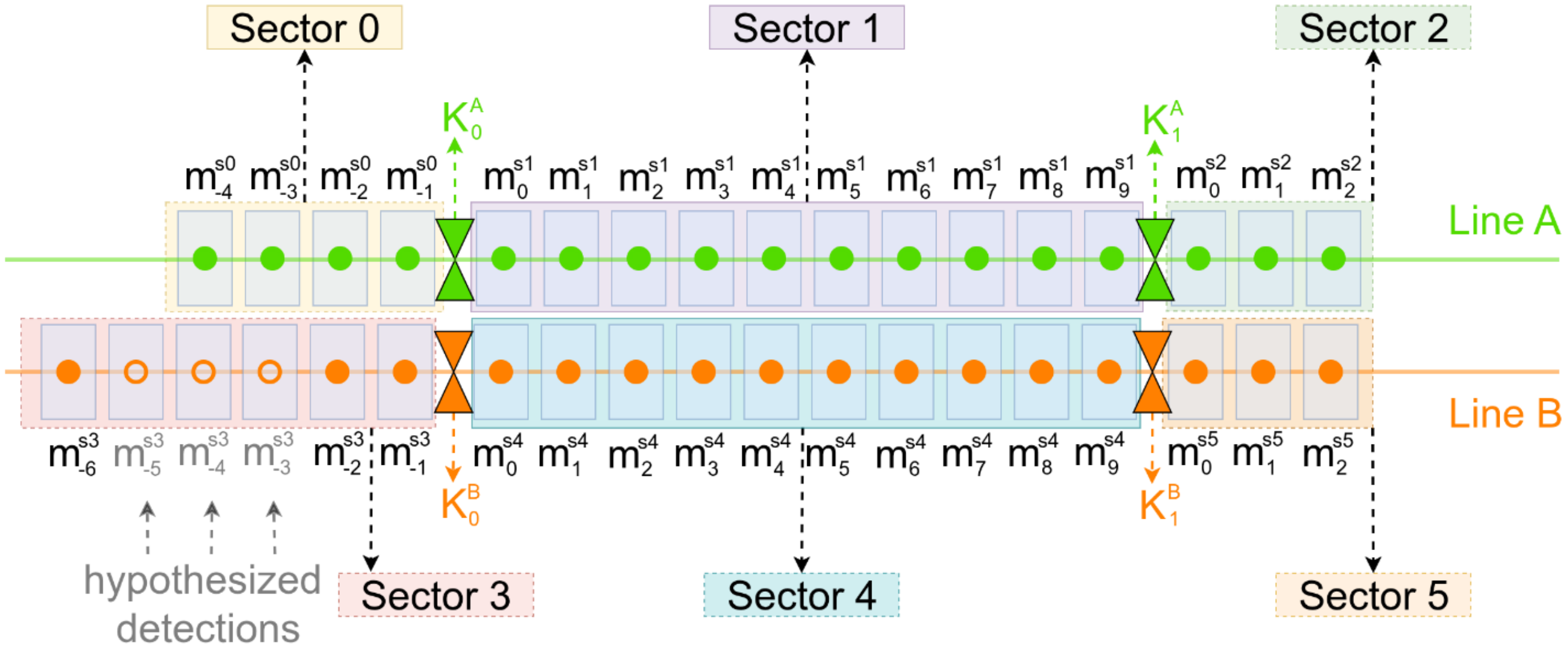

**Figure 9.** An illustration of an inferred power plant structure with PV module ($m_j^{si}$) and keypoint ($K_j^i$) detections.

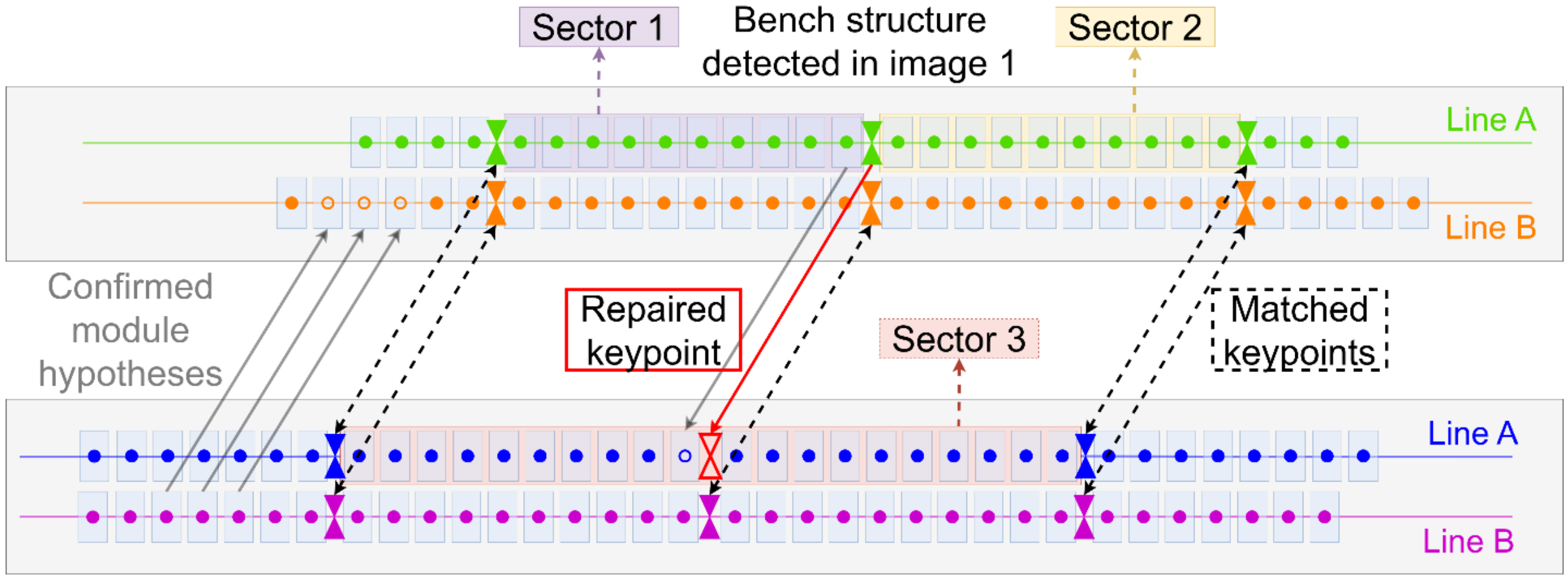

**Figure 10.** Matching of structures from two images. Matched keypoints are connected by dashed lines and confirmed module hypotheses are connected by gray lines. A missing keypoint detection illustrated in red is repaired in sector 3.

Afterwards, groups with common elements (sectors) are merged. A unique and globally consistent ID is assigned to all sectors that belong in the same group.

Following this, we verify the consistency of the sectors belonging to each group. The number of PV modules in all the matched sectors should be equal and each group should have sectors connected to keypoints from maximally two different keypoint groups. It is also possible to manually verify inconsistencies between individual images by displaying the respective images and detected structures.

An inconsistency may arise in the case of missing keypoint detections, resulting in multiple different sectors being categorized into the same sector group. This occasionally occurs when a PV module is not detected near the keypoint in question. An example of this is shown in Figure 10, where sectors one, two, and three would be incorrectly matched together. This can be repaired by inserting a new keypoint at a relevant position in the PV module sequence in the faulty sector number three, splitting it into two new sectors. Afterwards, the sectors of affected groups are updated with a new group ID.

The consistency of the number of modules in each sector of the same group is verified, and hypothesized modules are confirmed using detections from other images. Unconfirmed hypothesized modules are discarded. Finally, module detections are matched between sectors and each one is assigned a global module ID. The resulting globally matched structures are shown in Figure 11.

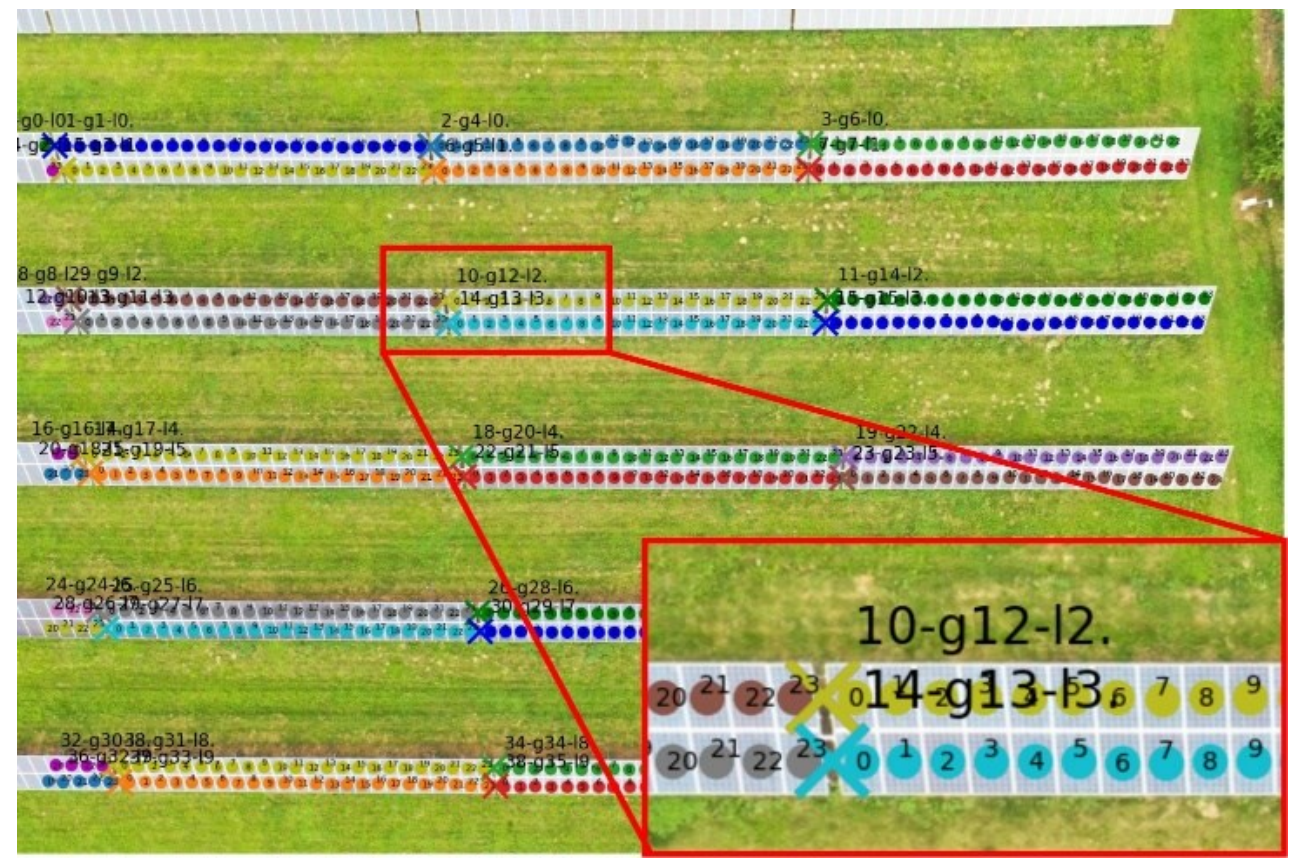

**Figure 11.** An image marked with global semantic structure information. Individual bench sectors are color-coded and marked with their unique ID, their global group ID, and their global line ID (e.g., 10-g12-l2). PV modules are marked with their position in the line sequence.

### 3.6. Optimization of the 3D model

The 3D positions of each PV module detection are calculated using the 3D model from the SfM-based reconstruction (as described previously for the keypoint 3D pose calculation). The positions of each module are averaged using 3D coordinates calculated from each image in which the module was detected. The normals are averaged in the same way.

The layout of the power plant generally follows standard conventions, allowing optional refinement of module positions. Individual benches are typically planar, and the modules are evenly spaced.

We use the RANSAC algorithm to fit a line model to the 3D coordinates of the modules from each row of the bench. The lines are defined by their origin $\mathbf{p_i} = (x_0, y_0, z_0)$ and a direction vector $\mathbf{d_i} = \langle d_x, d_y, d_z \rangle$. We average the line model parameters from each row to estimate the main axis of the bench (Figure 12).

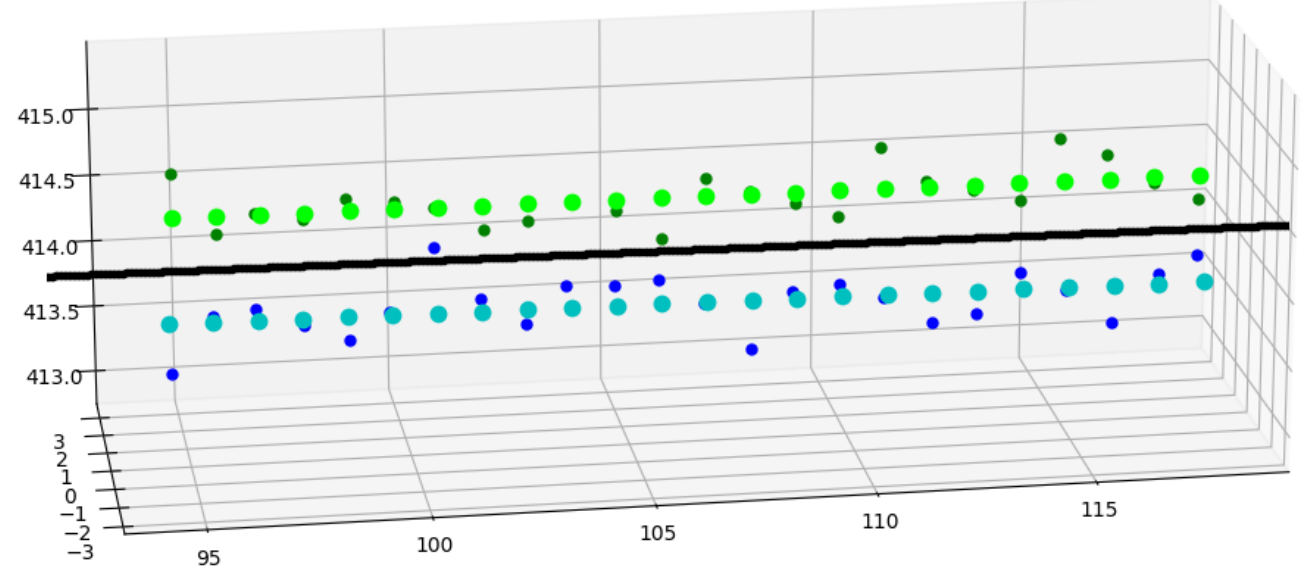

**Figure 12.** Optimization of PV module positions. Original positions are depicted in green and blue, new positions in light green and light blue. The axis of the bench is represented by a black line.

The module positions are orthogonally projected onto the main axis and used to compute direction vectors (Equation 4).

$$\mathbf{p}_{\perp} = \mathbf{p}_{bench} + [(\mathbf{p} - \mathbf{p}_{bench}) \cdot \mathbf{d}_{\text{bench}}]\, \mathbf{d}_{\text{bench}} \tag{4}$$

The direction vectors of the modules in each row are averaged to obtain a consistent orientation and distance for each row of PV modules. Using the averaged direction vectors, the modules are then repositioned along the row at uniform intervals.

The optimized 3D positions and normals of the modules, together with information on their assignment in the structure of the power plant, constitute the final model of the power plant.

## 4. Experiments and results

The developed mapping pipeline was tested on two photovoltaic power plants. The main characteristics of power plant 1 (PP1) and power plant 2 (PP2) are provided in Table 1. Detailed characteristics are described below.

Table 1. Dimensions, module and bench counts, and numbers of images used for detection and supplementary 3D reconstruction in each power plant.

| | Size [m] | N. of PV modules | N. of benches | N. of images (detection + supplementary) |
|---|---|---|---|---|
| PP1 | 275 x 140 | 4974 | 106 | 19 + 14 |
| PP2 | 160 x 110 | 5400 | 27 | 17 + 13 |

The first power plant (PP1) has dimensions of 275 × 140 meters (Figure 13). The power plant consists of 13 continuous lines with 106 PV benches, each bench consists of two parallel rows of PV modules (Figure 6a). The total number of modules in the power plant is 4974. The total number of sectors and bench gaps is 212 and 184, respectively. For the detection of PV modules, 19 images were used. In addition, 14 supplementary images from higher altitudes, ranging from 120 to 230 meters above ground, were included to improve the OpenSfM 3D reconstruction results.

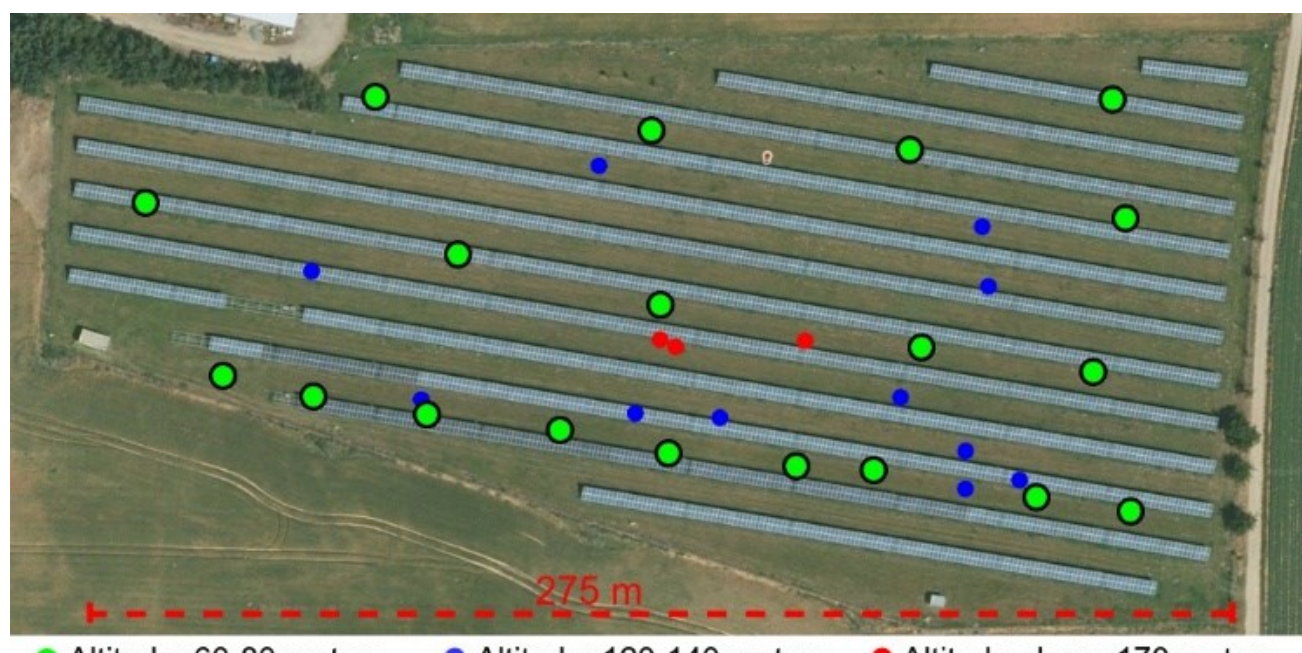

**Figure 13.** Aerial view of the PP1 power plant with marked image capture positions [18]. Positions of images used for PV module detection are accentuated by black circles.

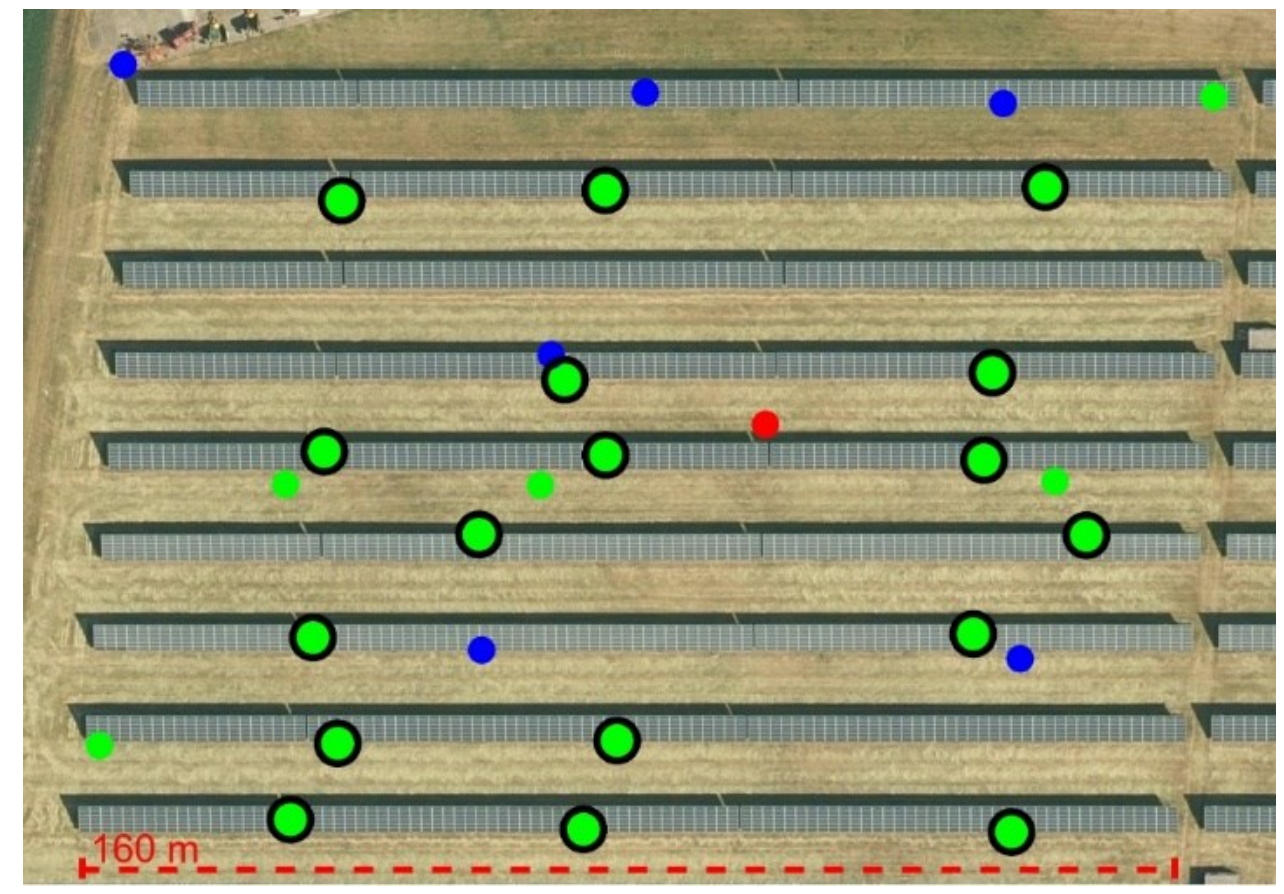

**Figure 14.** Aerial view of the PP2 power plant [18]. Positions of images used for PV module detection are accentuated by black circles.

The dimensions of the second power plant (PP2) are 160 × 110 meters (Figure 14). The power plant consists of 9 lines with 27 benches, each consisting of six parallel rows of PV modules (Figure 6b). It contains 5400 PV modules. The total number of sectors and bench gaps is 162 and 108, respectively. The detection of PV modules was performed using 17 images and 13 supplementary images were used to improve the 3D reconstruction.

The overview images for the PP1 power plant were captured using a Hasselblad camera capable of acquiring images with a resolution of 5472 × 3648 pixels. The images used for the proposed mapping framework were resized to half the resolution (2736 × 1824) to reduce the computational complexity of the 3D reconstruction and detection algorithms. The images for PP2 were captured with a DJI Zenmuse H20T camera used in wide mode. The images were used with their original resolution of 4056 × 3040 pixels.

The detection of the PV modules was performed using the YOLO and U-Net CNN models trained by the authors of [8] and described therein. The training dataset consisted of more than 3000 PV module annotations, which were used to train both the YOLO and the U-Net models. The dataset includes a mixture of aerial inspection images and publicly available images with different PV installations. The annotated dataset was split into training (80 %) and validation (20 %) partitions.

### 4.1. Processing times

The number of overview images and PV modules in both power plants is comparable. Manually capturing a complete set of images for one power plant takes approximately 8 minutes. In comparison, the flight duration reported in [12] for a power plant about half the size of PP1 was 30 minutes (the reported time did not include the subsequent image processing and 3D reconstruction).

The processing times in our pipeline are as follows. The 3D reconstruction using OpenSfM takes 10-15 minutes. The detection of PV modules and keypoints takes about 15 minutes (post-processing of YOLO detections accounts for roughly 60 % of this time). Calculating the 3D positions of all detections takes 10-15 minutes. The final fusion of detected structures across individual images and the optimization of the 3D model take less than one minute. The presented computation times include the generation of optional intermediate outputs (visualizations), which constitute approximately 15 % of the total processing time.

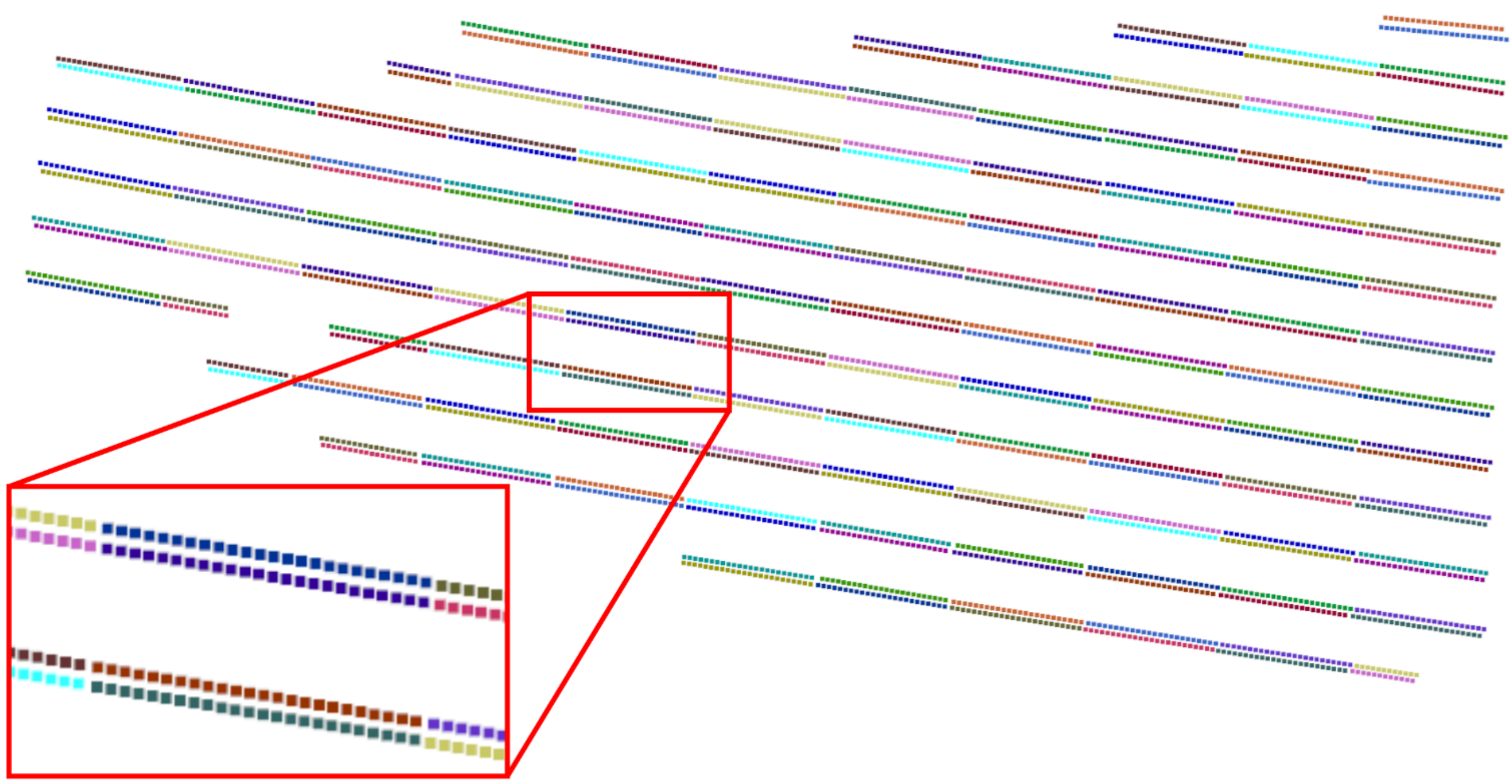

**Figure 15.** Compact 3D model of the PP1 power plant produced by the proposed mapping pipeline. Individual bench sectors are highlighted in different colors to distinguish inferred structural components.

The computation was performed on a computer with an Intel Core i7-7700 @ 3.60 GHz processor, 64 GB RAM, and an Nvidia GeForce 2080Ti GPU.

### 4.2. PV module detection

The average success rate for PV module detection across all images in both datasets exceeded 98.5 %. On average, each module and bench gap were detected in approximately 2.75 images.

The PV module detection success rates for the worst image were 97.8 % in PP1 and 92.9 % in PP2. From the missing detections 68 % was caused by strong sun reflection. The maximum number of hypothesized missing detections did not exceed three in any image. The missing detections were generally registered in other images.

In PP1, the detections were derived mainly from U-Net, which contributed 80.1 %, while the YOLO detections comprised the remaining 19.9 %. The total number of PV module detections was 13322. The number of detections after the RANSAC filtering and alignment was 13240.

In PP2, the detections from U-Net contributed 90.7 % and the YOLO detections comprised the remaining 9.3 %. The total number of PV module detections was 19973 before the RANSAC filtering and 19910 after.

The success rates for the detection of bench gaps (keypoints) were 93.5 % and 91.48 %, for PP1 and PP2 respectively. Of these undetected bench gaps, 38.3 % were caused by missing detections of their neighboring PV modules; the rest is due to incorrect classification. The total number of detected bench gaps was 494 for PP1 and 458 for PP2 (approximately 26 per image).

The mapping framework successfully registered and mapped all 5400 PV modules in the PP2 power plant without any need for manual intervention. For the PP1 power plant, only 4971 from the total number of 4974 PV modules were mapped. The undetected modules were identified by an uneven number of modules in their respective benches. The modules were present only in a single image and had to be marked manually.

### 4.3. 3D model

The result is a compact georeferenced model that defines the structure of the power plant and the exact position, orientation, and association of the individual photovoltaic modules. The model of the PP1 power plant can be seen in Figure 15.

#### 4.3.1. Internal characteristics of the 3D model

The internal consistency of the 3D models was evaluated. Figure 16 displays the differences between the position of the modules before and after the optimization procedure for both power plants.

While such deviations may have a negligible impact on coarse inspection procedures, they can cause irregularities in fine-grained inspection tasks that rely on precise per-module positioning. Some autonomous vision-based inspection systems even require a precise 3D model of the power plant for localization (e.g., [8]).

The difference values also illustrate the level of precision of the raw SfM-based 3D reconstruction. Inconsistencies are most pronounced along the z-axis, which mainly coincides with the z-axis of the camera. This is a common limitation of SfM-based reconstructions, where depth estimation tends to be less reliable than planar positioning. The semantic structures detected directly in the images exhibit minimal imprecision. This ensures their consistency and allows the optimization procedures to effectively mitigate the inaccuracies in the raw 3D model.

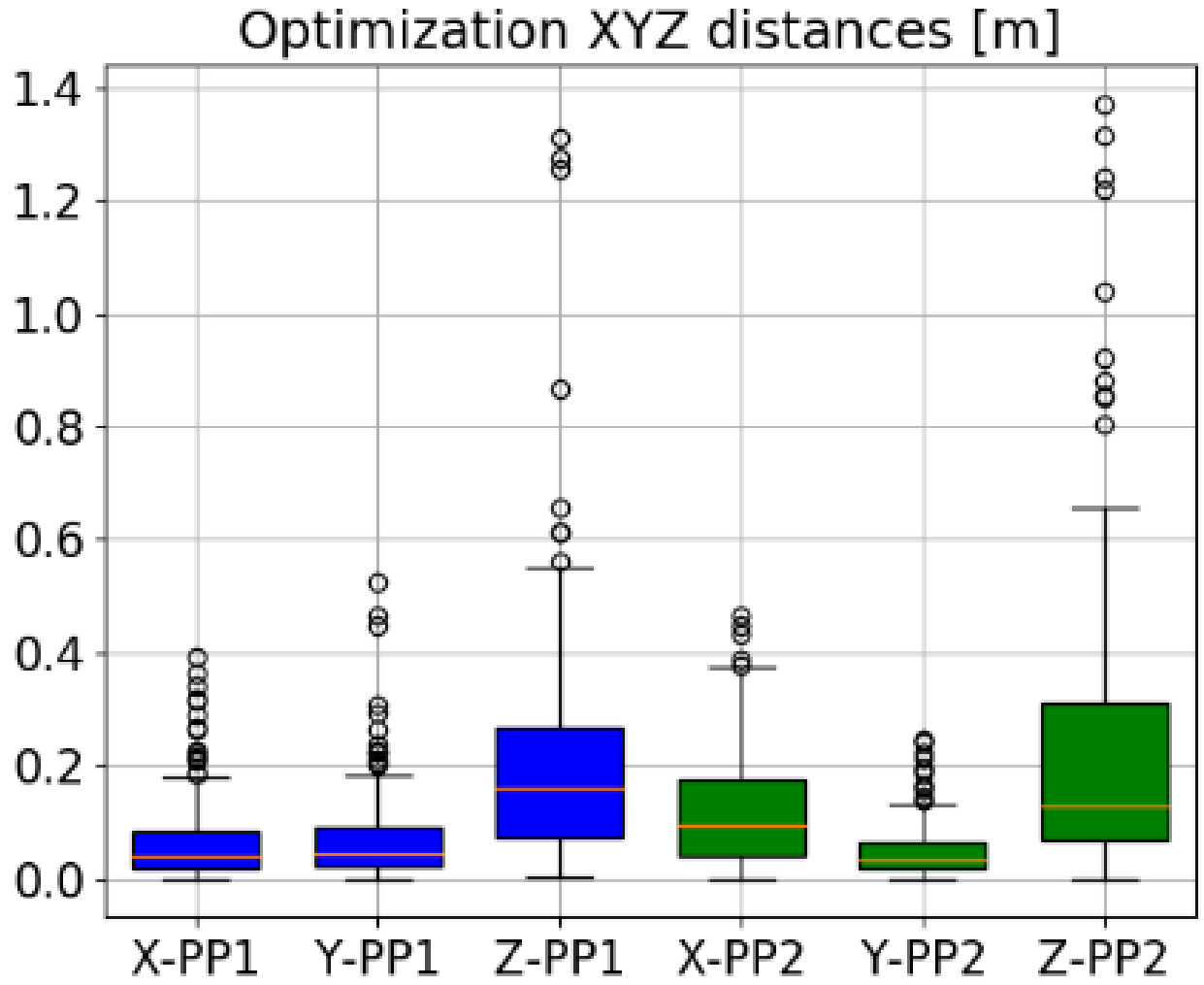

**Figure 16.** Five point statistics of the x, y, and z position differences before and after the optimization procedure.

The dimensions of the PV modules in PP1 are 1.0 × 1.65 meters. For PP2, these dimensions are 1.57 × 0.79 meters. Figure 17 displays the distances between neighboring module centers after optimization. We can see that the distances in the mapped model reflect the expected spacing in the PV installations with reasonable accuracy. Optimal distance values can be enforced during the optimization procedure if required.

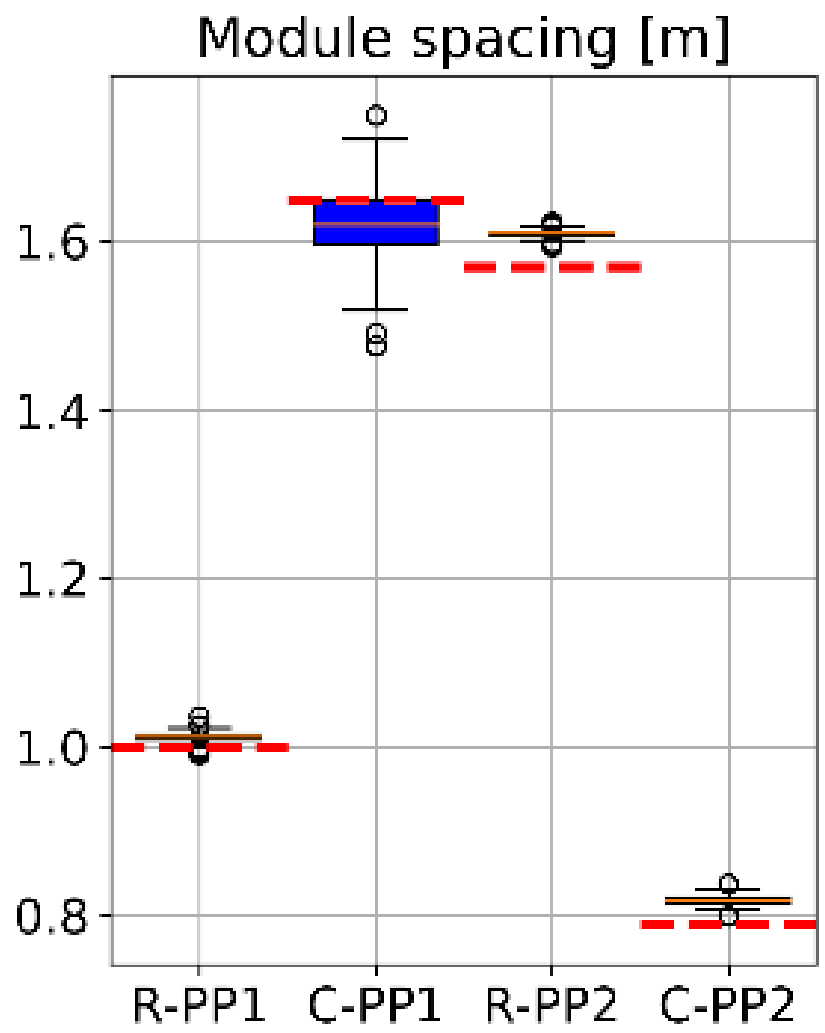

**Figure 17.** Five point statistics of distances between individual PV module centers in rows (R) and columns (C). The respective PV module dimensions are depicted for reference using dashed red lines.

### 4.3.2. Accuracy of the 3D model

The topology of the final layout is consistent with the technical drawings provided by the owner of each power plant. We further evaluate the accuracy of the generated 3D models by comparing them to reference models.

Each reference model was created manually, using publicly available orthophoto [18] and elevation maps [19]. The layout of the power plant was inferred from the orthophotos and several PV modules at the edges of the power plant were manually georeferenced to align the model with the global coordinate system. The model was combined with data from elevation maps to ensure consistency with ground level. Lastly, the modules were elevated from the ground and oriented accordingly to on-site measurements of the height and tilt of the benches. The margin of error in the reference model is estimated to be around 0.5 meters.

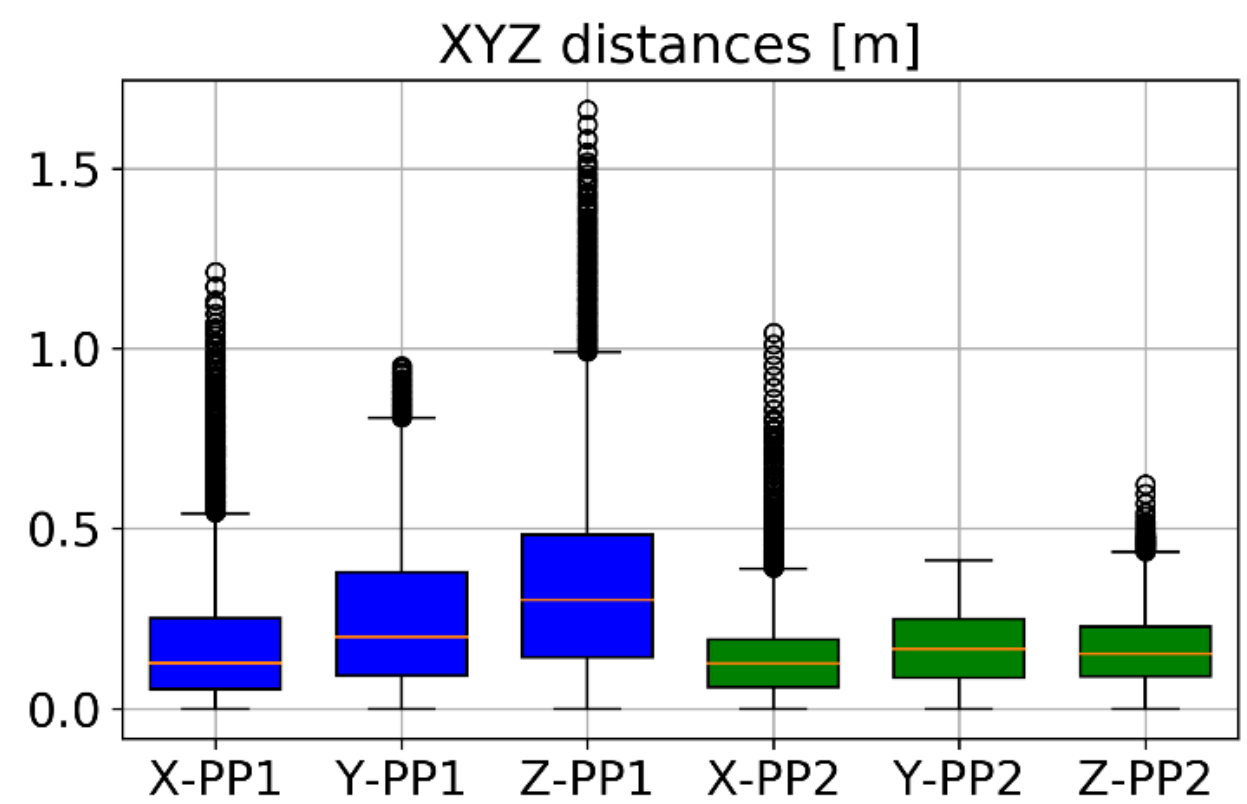

**Figure 18.** Five point statistics of position differences between our 3D models and the manually created reference models.

Individual modules from our 3D models were paired with modules from the reference models using the topological layout of the power plant. The manually georeferenced modules were also used to align our models with the global coordinate system. Figure 18 shows the deviations between the module positions in the two models. Since the models are internally consistent (as shown in the previous section), the position deviations can be attributed to relative shifts between the positions of individual benches.

### 4.4. Discussion and future work

The topological consistency of the created models, along with their internal precision, and global accuracy make the resulting 3D models suitable for power plant maintenance and most autonomous inspection procedures.

Maintenance systems can directly benefit from the knowledge of the power plant structure produced by our method. Direct assignment of defective modules to known segments can be used to plan more detailed flights above segments with known defects. The positive benefit-cost ratio of such approaches is discussed in several works [20, 21]. The structure also often coincides with the wiring (electrical clustering) of the PV modules in series and parallel [2].

An absolute error in the range of 1-2 meters is sufficient for global path planning and allows observed keypoints (e.g., bench borders) to be correctly associated with the corresponding bench in the model during inspection. Meanwhile, an internal (relative) positioning error on the order of centimeters is sufficient to maintain reliable module association while flying along a bench. Additionally, in thermal-based anomaly detection, the data must be acquired under an optimal viewing angle and distance [22]. Models generated by the proposed method could be utilized in various works on photovoltaic power plant inspection [2, 3, 4, 8, 23].

The images in this work were captured manually by a skilled drone operator. In future work, we would like to automate the data acquisition process. A flight plan defined by GNSS waypoints, ensuring sufficient overlap and coverage of the power plant, can provide the input required for the mapping pipeline in automated operations.

The presented method is based on design concepts commonly found in a large number of PV power plants and its reliance on visually identifiable keypoints may limit its applicability to more specific layouts. It is possible to solve some generalizability issues, e.g., we can assume that PV module rows project into curved lines when installed on uneven terrain. However, given the diversity of irregular layout configurations, we believe that more specific modifications should be designed on a case-by-case basis.

## 5. Conclusion

This work presents an automated mapping pipeline for the creation of power plant models. The process produces a detailed model down to the level of individual PV modules. This enables us to infer an up-to-date layout of the power plant. Thus, removing the reliance on third-party data, which may be unreliable or become outdated over time.

The approach relies on visual segmentation of PV modules and inference of structural information in overview images. The structure-from-motion technique is used to reconstruct a 3D model of the power plant. However, we acknowledge the inherent limitation of 3D reconstruction accuracy, which makes direct merging of PV module detections between individual images unfeasible. To address this, we designed a set of visually identifiable keypoints related to the structural layout and used these to merge the power plant structures between individual images.

To account for possible limitations of the PV module segmentation methods, the pipeline also allows optional manual verification and intervention at several stages.

Experimental validation attests to the practicality and effectiveness of the proposed approach. Two power plants with areas of 275 × 140 and 160 × 110 meters were mapped using only 33 and 30 overview images, respectively. The internal quality of the created georeferenced models was evaluated, and the accuracy of these models was compared to manually created 3D models. The proposed mapping pipeline achieved satisfactory results. The created power plant models are suitable for navigation, inspection, and maintenance.

## Acknowledgement

This work was co-funded by the European Union under the project ROBOPROX (reg. no. CZ.02.01.01/22_008/0004590).

## Author contributions

**Viktor Kozák:** Conceptualization, Methodology, Software, Validation, Investigation, Writing – Original Draft, Review & Editing, Visualization. **Jan Chudoba:** Software Resources, Data curation, Writing – Review & Editing. **Libor Přeučil:** Supervision, Funding acquisition.

## Conflicts of interest

The authors declare no conflicts of interest.